\documentclass[a4paper,fleqn]{cas-dc}

\usepackage[numbers]{natbib}
\usepackage{subcaption}
\newcommand{\xmark}{\ding{56}}%
\usepackage{pifont}

\usepackage{lineno}

\def\tsc#1{\csdef{#1}{\textsc{\lowercase{#1}}\xspace}}
\tsc{WGM}
\tsc{QE}
\tsc{EP}
\tsc{PMS}
\tsc{BEC}
\tsc{DE}

\begin{document}


\let\WriteBookmarks\relax
\def\floatpagepagefraction{1}
\def\textpagefraction{.001}
\shorttitle{FA-Seg}
\shortauthors{Q. -H. Che et~al.}

\title [mode = title]{\textbf{FA-Seg}: A Fast and Accurate Diffusion-Based Method for Open-Vocabulary Segmentation}

\author[1,2]{Huy Che}[orcid=0009-0007-7477-4702]
\ead{huycq@uit.edu.vn}
\credit{Writing – original draft, Visualization, Methodology, Investigation, Conceptualization, Software}

\author[1,2]{Vinh-Tiep Nguyen}[orcid=0000-0003-4260-7874]
\ead{tiepnv@uit.edu.vn}
\cormark[1]

\credit{Writing – review and editing, Supervision, Conceptualization}

\affiliation[1]{organization={University of Information Technology},
                city={Ho Chi Minh City},
                country={Vietnam}}
                
\affiliation[2]{organization={Vietnam National University},
                city={Ho Chi Minh City},
                country={Vietnam}}

\cortext[cor1]{Corresponding author}

\begin{abstract}
Open-vocabulary semantic segmentation (OVSS) aims to segment objects from arbitrary text categories without requiring densely annotated datasets. Although contrastive learning based models enable zero-shot segmentation, they often lose fine spatial precision at the pixel level, due to global representation bias. In contrast, diffusion-based models naturally encode fine-grained spatial features via attention mechanisms that capture both global context and local details. However, they often face challenges in balancing the computation costs and the quality of the segmentation mask. In this work, we present FA-Seg, a \textbf{F}ast and \textbf{A}ccurate training-free framework for open-vocabulary segmentation based on diffusion models. FA-Seg performs segmentation using only a \textit{(1+1)-step} from a pretrained diffusion model. Moreover, instead of running multiple times for different classes, FA-Seg performs segmentation for all classes at once. To further enhance the segmentation quality, FA-Seg introduces three key components: (i) a \textit{dual-prompt mechanism} for discriminative, class-aware attention extraction, (ii) a \textit{Hierarchical Attention Refinement Method (HARD)} that enhances semantic precision via multi-resolution attention fusion, and (iii) a \textit{Test-Time Flipping (TTF)} scheme designed to improve spatial consistency. Extensive experiments show that FA-Seg achieves state-of-the-art training-free performance, achieving 43.8\% average mIoU across PASCAL VOC, PASCAL Context, and COCO Object benchmarks while maintaining superior inference efficiency. Our results demonstrate that FA-Seg provides a strong foundation for extendability, bridging the gap between segmentation quality and inference efficiency. The source code is available at \url{https://github.com/chequanghuy/FA-Seg}.
\end{abstract}

\begin{keywords}
Open-vocabulary segmentation \sep Diffusion models \sep Training-free \sep Attention Refinement
\end{keywords}

\maketitle

\section{Introduction}

Traditional semantic segmentation models rely heavily on fully supervised training using densely annotated datasets. While effective within a fixed set of predefined categories, these models struggle to scale to the vast diversity of visual concepts encountered in real-world settings due to labor-intensive annotated datasets. To address this limitation, Open-Vocabulary Semantic Segmentation (OVSS) has emerged as a compelling alternative, aiming to segment images based on arbitrary textual inputs without requiring manual annotations for every possible class. Recent studies have extended open-vocabulary recognition into multi-view 3D representations by embedding language-aligned features into Neural Radiance Fields (NeRFs) \cite{LERF, LASER, NeRF}. These NeRF-based methods enable concept localization and querying within reconstructed 3D radiance fields, focusing primarily on scene-level exploration and spatial reasoning. In contrast, open-vocabulary semantic segmentation (OVSS) operates in the 2D pixel space, where the goal is to assign semantic labels to every pixel within a single image. Our work follows this pixel-level direction, achieving open-vocabulary segmentation in natural images without the complex multi-view data preparation and training pipelines required by NeRF-based approaches.

\begin{figure*}[t]
    \centering
    \begin{subfigure}[b]{0.37\textwidth}
        \includegraphics[width=\textwidth]{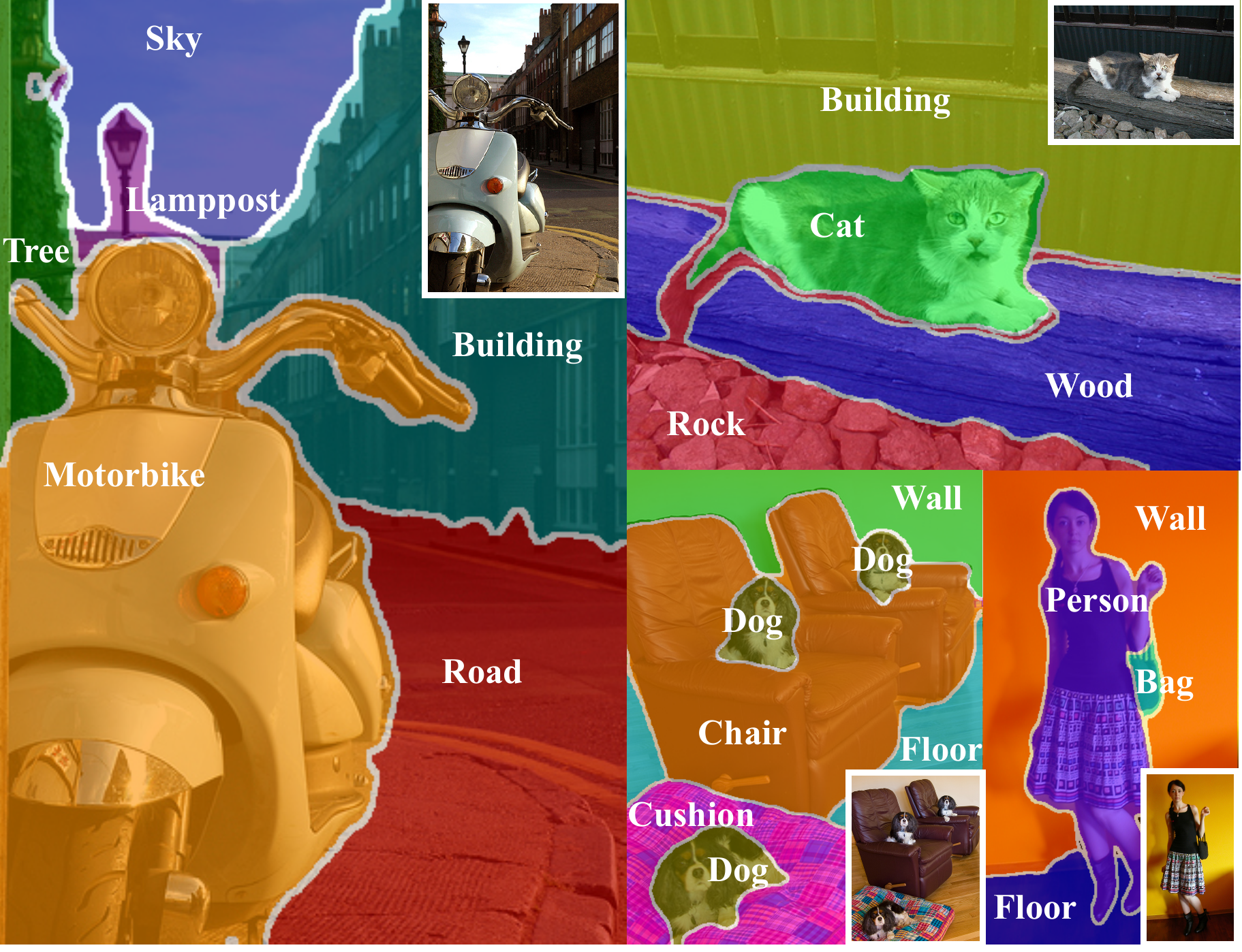}
    \end{subfigure}
    \begin{subfigure}[b]{0.325\textwidth}
        \includegraphics[width=\textwidth]{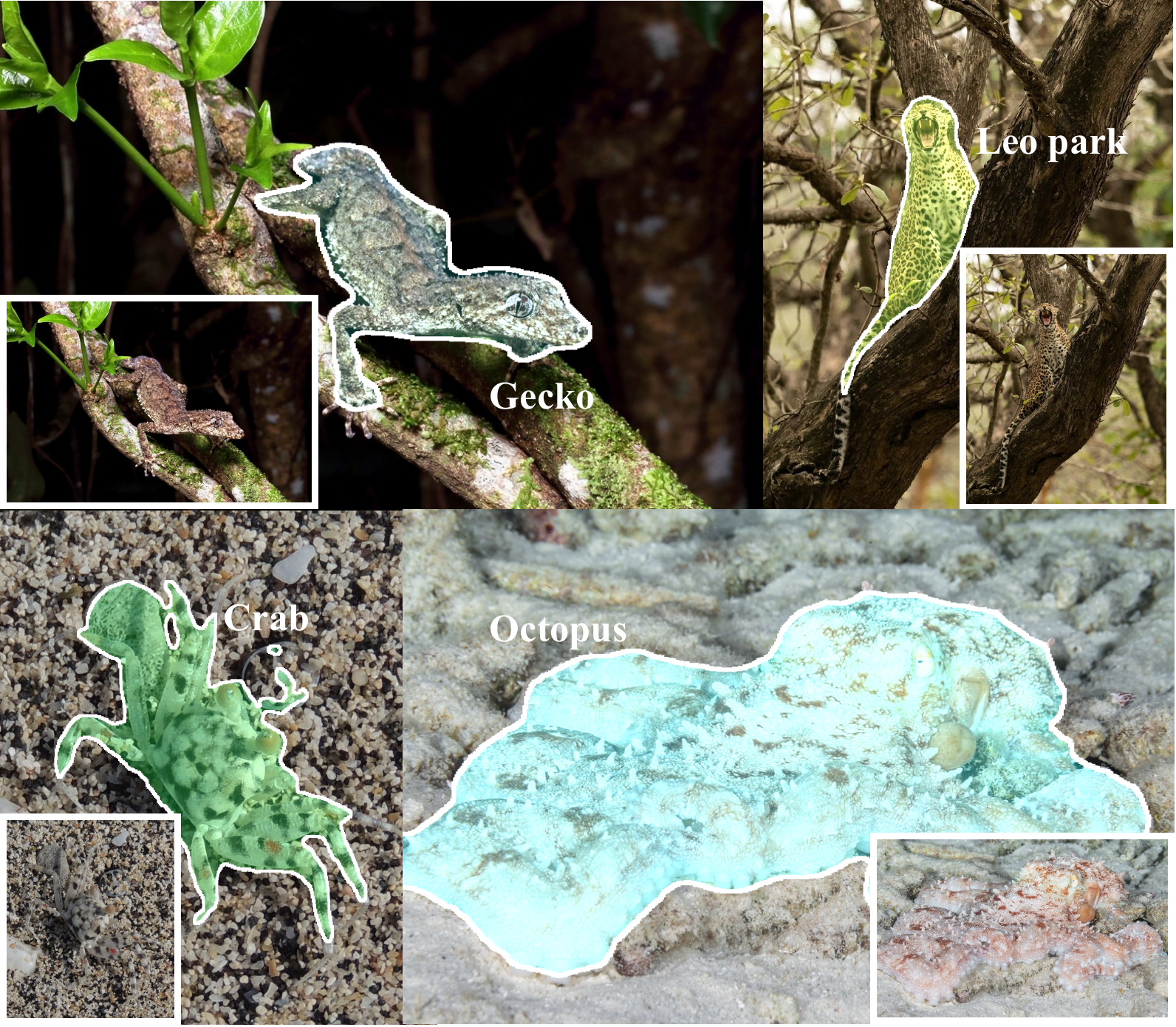}
    \end{subfigure}
    \begin{subfigure}[b]{0.27\textwidth}
        \includegraphics[width=\textwidth]{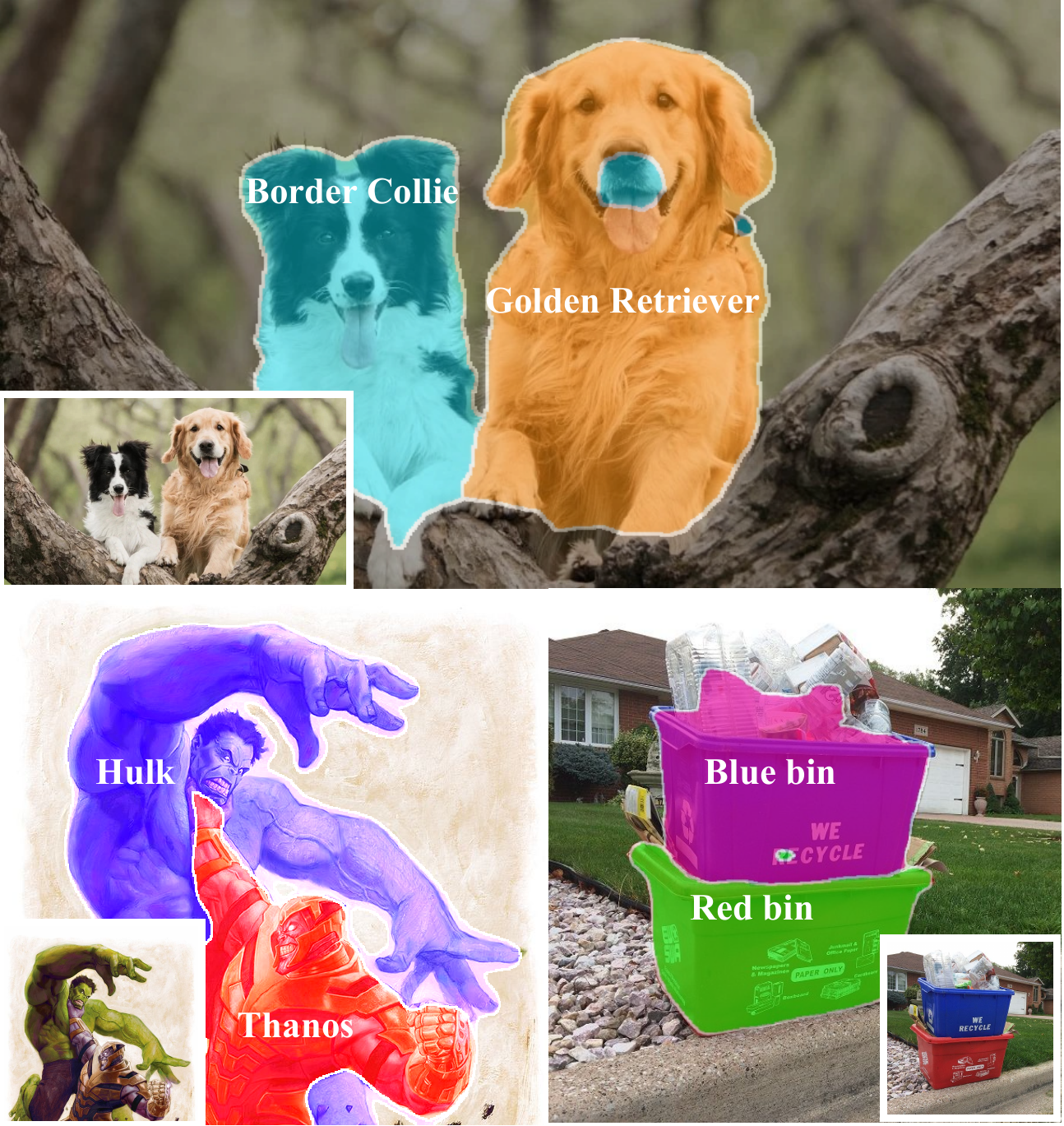}
    \end{subfigure}
    \caption{\textbf{Open-vocabulary semantic segmentation results using FA-Seg.} We visualize predictions on challenging scenarios, including contextual images, camouflaged objects, and internet-collected images. The highlighted results demonstrate that, in addition to accurate segmentation of both stuff and things, our method effectively handles camouflaged objects, distinguishes between different characters or dog breeds, and differentiates objects based on color variations.}
    \label{fig:highlight}
\end{figure*}

Existing OVSS methods primarily follow two dominant paradigms: vision-language models and generative models. Vision-language models such as CLIP \cite{clip} leverage contrastive learning to align visual and textual representations within a shared embedding space. Approaches based on CLIP can be broadly categorized into training-free methods \cite{reco, CLIP-DIY,sclip,clearclip}, and training methods \cite{tcl,ZeroSeg,viewco,SegCLIP,ovsegmentor,simseg}. However, these models often suffer from limited spatial precision due to their reliance on global image-text embeddings. In contrast, generative models \cite{odise, SegLD, ovdiff, freeda, FreeSegDiff, diffsegmenter, InvSeg} have recently emerged as a powerful alternative, as their generative attention mechanisms inherently preserve spatial structures, enabling them to capture both object location and shape with high fidelity. Nevertheless, many diffusion-based methods require extensive supervised training and computationally expensive synthetic dataset generation, limiting their scalability and practical deployment.

To address these challenges, recent research has explored training-free diffusion-based OVSS approaches that eliminate the need for supervision or retraining. While promising, existing methods often trade inference efficiency for segmentation quality: some require repeated inference for each candidate class, while others involve time-consuming optimization or prompt tuning steps. This creates a critical gap between segmentation accuracy and computational efficiency, particularly for real-world open-vocabulary scenarios. These limitations reveal a critical open question: \textit{How can training-free diffusion-based OVSS methods achieve high segmentation quality while remaining computationally efficient enough for practical deployment?}


In this paper, we present \textbf{FA-Seg}, an efficient training-free framework for open-vocabulary semantic segmentation (OVSS) that leverages pretrained text-to-image diffusion models, as illustrated in Figure~\ref{fig:highlight}. FA-Seg reconstructs the input image using fast (1+1)-step DDIM inversion. The dual-prompt mechanism is used in addition to class-aware semantic enhancement, allowing for simultaneous multi-class segmentation without iterative inference. In addition to Test-Time Flipping (TTF), to further improve segmentation quality, we propose a multi-scale attention refinement strategy, \textbf{HARD} (\textbf{H}ierarchical \textbf{A}ttention \textbf{R}efinement Metho\textbf{d}), which fuses cross-attention and self-attention maps across resolutions to enhance object localization and boundary accuracy. We summarize our key contributions as follows:

\begin{itemize}
    \item We introduce FA-Seg that leverages diffusion models with fast (1+1)-step DDIM inversion and a unified class prompt, enabling efficient attention map generation for all candidate classes in a single inference run.

    \item To enhance segmentation accuracy, we propose three key techniques: (i) a \textit{dual-prompt mechanism} that separately guides semantic reconstruction and class-specific attention extraction to help better class awareness, (ii) the proposed \textit{HARD} improves fused cross-attention maps using transformed self-attention maps at multi-scales, and (iii) \textit{TTF} that improves the spatial consistency of the segmentation mask.

    \item We conduct comprehensive experiments on three benchmark datasets-PASCAL VOC, PASCAL Context, and COCO Object-demonstrating that FA-Seg achieves state-of-the-art performance among training-free methods while maintaining superior computational efficiency.

\end{itemize}

\section{Related work}

\subsection{Contrastive Vision-Language Models for OVSS}

While self-supervised learning methods such as MoCo \cite{moco} and DINO \cite{dino} have shown success in learning visual features, they struggle with flexible semantic understanding for unseen categories. Large-scale vision–language models like CLIP \cite{clip} address this limitation by aligning visual and textual features in a shared embedding space, making them suitable for open-vocabulary semantic segmentation (OVSS).

Several recent \textbf{training-free} methods directly exploit CLIP’s inherent localization capability by extracting or refining attention maps for segmentation. MaskCLIP \cite{maskclip} utilizes value embeddings from CLIP’s final self-attention layer for label-free segmentation. SCLIP \cite{sclip} and CLIPSurgery \cite{CLIPSurgery} modify the query-key attention mechanism via self-self attention to better focus on semantic regions, while ClearCLIP \cite{clearclip} simplifies CLIP’s final transformer layer to improve fine-grained localization. Other works, such as ReCo \cite{reco}, leverage CLIP’s retrieval capability and the alignment between visual features and category semantics. Additionally, CLIP-DIY \cite{CLIP-DIY} combines unsupervised object localization with CLIP to generate foreground-background guidance maps.

In contrast, some \textbf{unsupervised training} approaches learn segmentation models from image–text pairs without relying on pixel-level labels. GroupViT \cite{GroupViT}, SegCLIP \cite{SegCLIP}, and OVSegmentor \cite{ovsegmentor} introduce group tokens into the model backbone to learn semantic clusters aligned with textual concepts. Specifically, GroupViT treats group tokens as class proxies from text supervision, SegCLIP combines CLIP features with learnable tokens, while OVSegmentor introduces masked entity completion and cross-image mask consistency to improve segmentation. ZeroSeg \cite{ZeroSeg} distills CLIP's visual concepts into learnable segmentation tokens for localizing semantic regions, and ViewCo \cite{viewco} enforces multi-view consistency across augmented views. TCL \cite{tcl} leverages masked regions to improve semantic alignment between images and text. Although \textbf{fully supervised} methods \cite{fc-clip,SAN} that utilize dense segmentation annotations typically achieve higher performance than unsupervised or training-free methods, they are less favored in the context of open-vocabulary semantic segmentation due to the high cost of annotation and the complexity of training on large-scale datasets.

In this study, we propose a training-free open-vocabulary semantic segmentation (OVSS) approach, driven by the goal of reducing annotation overhead and enhancing scalability-while maintaining flexibility in segmenting diverse visual concepts. Unlike prior methods that predominantly rely on vision-language models with contrastive alignment, which often compromise spatial precision, our approach adopts a generative pre-trained model. By inherently preserving the spatial structure of diffusion models through attention maps, we enable more accurate and semantically grounded segmentation without the need for additional training or pixel-level annotations.

\subsection{Semantic Segmentation with Diffusion Models}


Recently, diffusion models such as Stable Diffusion have shown strong potential in visual localization due to their ability to align textual prompts with spatial layouts of objects via attention maps. This enables them to capture both the position and shape of objects, opening up a promising direction for semantic segmentation in general, and open-vocabulary semantic segmentation in particular, without the need for manual annotations.


Several works \cite{attn2mask,diffumask,datasetdiff} utilize diffusion models to generate synthetic semantic segmentation datasets by extracting cross-attention maps from generated images. However, these methods are limited to synthetic images and require additional segmentation models to handle real-world data. To mitigate this issue, recent studies \cite{diffseg_un,diffcut} apply unsupervised zero-shot segmentation directly to real images by leveraging self-attention maps from pre-trained diffusion models. DiffSeg \cite{diffseg_un} use clustering to group semantic regions, while DiffCut \cite{diffcut} applies graph-based segmentation for better boundary separation. Nonetheless, lacking cross-attention, these methods cannot assign semantic labels, limiting their applicability in open-vocabulary scenarios.

\begin{figure*}[t]
	\centering 
	\includegraphics[width=1.065\textwidth]{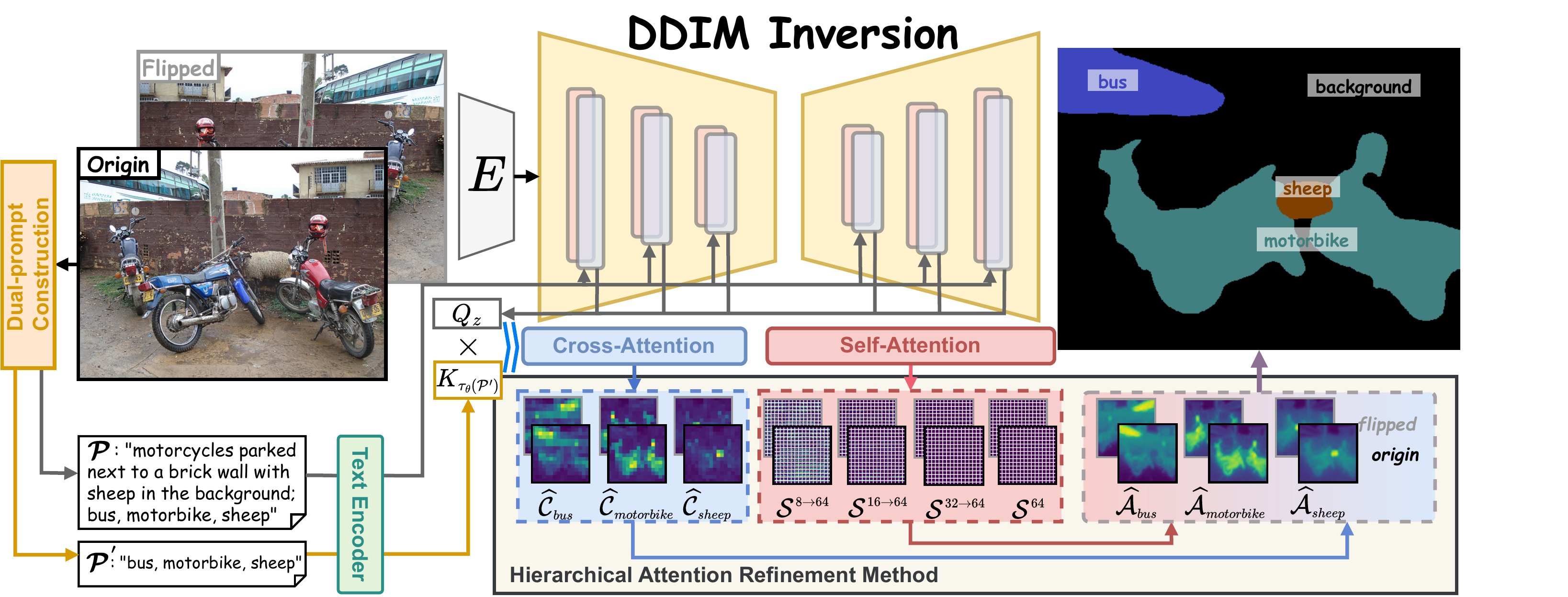}	
	\caption{\textbf{Overview of the proposed FA-Seg}. The input consists of both the original and horizontally flipped images, which are first reconstructed via DDIM inversion to obtain their latent representations. In parallel, a text prompt $\mathcal{P}$ and a class prompt $\mathcal{P'}$ are generated to guide inversion and extract class-specific cross-attention maps corresponding to candidate classes (e.g., bus, motorbike, sheep). Cross-attention and self-attention maps at multiple resolutions are then extracted, transformed, and fused via weighted aggregation, where the self-attention maps are used to refine the class-specific cross-attention maps. Final segmentation masks are derived from the refined per-class score maps. Test-Time Flipping (TTF) is applied during inference to enhance robustness: attention maps from the flipped image are spatially realigned and averaged with those from the original image to improve prediction reliability.} 
	\label{fig:main_pipeline}%
\end{figure*}

More recently, diffusion-based open-vocabulary segmentation has emerged as an alternative to CLIP-based methods. \textbf{Fully supervised} approaches \cite{odise,SegLD} achieve strong performance but require large annotated datasets and significant computational resources due to the heavy training demands of diffusion models. To overcome this, methods such as OVDiff \cite{ovdiff} and FreeDA \cite{freeda} propose generating \textbf{synthetic support sets} for query-to-support matching. While avoiding large-scale supervised training, these support-based methods involve complex data generation pipelines and struggle to generalize beyond fixed vocabularies. In contrast, \textbf{training-free} approaches are gaining increasing popularity due to their competitive performance without requiring model retraining. FreeSeg-Diff \cite{FreeSegDiff} clusters self-attention maps to generate class-agnostic object masks, which are then labeled by matching their embeddings with text embeddings via CLIP. Differing from this strategy, DiffSegmentor \cite{diffsegmenter} and InvSeg \cite{InvSeg} refine cross-attention using self-attention to localize semantically relevant regions and further enhance these representations to achieve more accurate segmentation. Although diffusion models generally incur higher computational costs compared to CLIP-based approaches, existing training-free methods primarily focus on improving segmentation performance while ignoring limited attention to optimizing inference efficiency.

In this work, we introduce a novel training-free open-vocabulary segmentation framework based on Stable Diffusion that simultaneously addresses the two key challenges of segmentation accuracy and inference efficiency. Unlike existing methods, our approach fully exploits both self-attention and cross-attention across multiple resolutions to capture rich semantic and spatial information. This enables the model to generate segmentation maps for all candidate classes in a single inference run achieving accurate, fine-grained segmentation while significantly improving computational efficiency, making it highly practical for real-world open-vocabulary applications.

\section{Methods}

\subsection{FA-Seg Overview}

Our objective is to perform training-free open-vocabulary semantic segmentation without requiring dense pixel-level annotations. Given a real image $I_0$ and a candidate class set, our proposed FA-Seg framework extracts segmentation masks via cross-attention and self-attention analysis from a pre-trained Stable Diffusion (SD) model. The overall framework is depicted in Figure~\ref{fig:main_pipeline}. The entire pipeline consists of three stages:

\begin{itemize}
    \item \textbf{Stage 1: Dual-Prompt Construction} (Sec.~\ref{sec:prompt}): We first construct two complementary text prompts: the caption prompt $\mathcal{P}$, and the class prompt $\mathcal{P}'$ containing the candidate categories.

    \item \textbf{Stage 2: Attention Extraction via Efficient Inversion} (Sec.~\ref{sec:inversion}): We employ a fast DDIM inversion using diffusion model to reconstruct the input image while extracting both cross-attention and self-attention maps across multiple resolutions $r \in \mathcal{R}$.

    \item \textbf{Stage 3: Hierarchical Attention Refinement} (Sec.~\ref{sec:refinement}): To produce accurate segmentation, cross-attention maps across multiple resolutions are fused into $\widehat{\mathcal{C}}_c$ via weighted aggregation. The fused cross-attention is then further refined using affinity matrices derived from self-attention across resolutions.
\end{itemize}

\subsection{Dual-Prompt Construction} \label{sec:prompt}

\begin{figure}[]
	\centering 
	\includegraphics[width=0.5\textwidth]{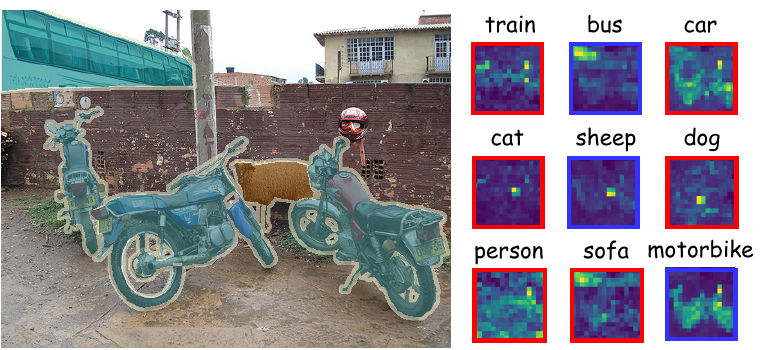}	
	\caption{\textbf{Qualitative results for the classes in the class prompt $\mathcal{P}'$.} Instead of including only the candidate classes that appear in the image (highlighted with \textcolor{blue}{blue bounding boxes}), we also include distractor classes (highlighted with \textcolor{red}{red bounding boxes}). The results show that while the candidate classes present in the image are visualized accurately, the distractor classes yield incorrect or noisy attention maps.} 
	\label{fig:all_class}%
\end{figure}

To enable semantic reconstruction and class-aware attention extraction, FA-Seg introduces a dual-prompt mechanism composed of $\Bigl(\mathcal{P}$, $\mathcal{P}'\Bigr)$. The textual prompt $\mathcal{P}$, generated by an image captioning model (e.g., BLIP), serves as the text condition for image reconstruction via DDIM inversion, facilitating faithful image generation. To explicitly inform the model about candidate classes, we append the class prompt $\mathcal{P}'$ to the textual prompt, following the structure: \texttt{"{image caption}; {$\mathcal{P}'$}"}. This design enables the diffusion model to focus on class-specific semantics during generation. The class prompt $\mathcal{P}'$ is constructed from a candidate set $\mathcal{C} = \{c_1, \ldots, c_M\}$. For example, if $\mathcal{C}$ = \{ \texttt{tv monitor, dog, person} \}, then $\mathcal{P}'=$ ``$tv$ $monitor,$ $dog,$ $person$". These candidate labels are essential for computing class-specific cross-attention maps that connect textual concepts to spatial regions in the semantic reconstruction of the image. Unlike contrastive models such as CLIP, which inherently suppress irrelevant concepts while enhancing relevant ones, text-to-image models like Stable Diffusion (SD) do not automatically ignore out-of-context objects. Instead, SD can preserve unrelated visual concepts as semantic noise. This behavior is illustrated in Figure~\ref{fig:all_class}, where distractor classes (those not present in the image) still produce activated attention maps. This motivates the need for a tailored class candidate set $\mathcal{C}$ for each image. 

To build the candidate set $\mathcal{C}$, we follow prior works \cite{diffsegmenter,InvSeg}, which combine BLIP and CLIP to generate candidate classes from a predefined vocabulary. To improve recall and label diversity, we replace CLIP with TagCLIP, a CLIP-based model optimized for multi-label classification. The TagCLIP+BLIP combination allows for more accurate and comprehensive identification of relevant object categories, improving the reliability of the candidate set $\mathcal{C}$.

To construct the candidate set $\mathcal{C}$, we adopt the approach from previous works \cite{diffsegmenter,InvSeg}, which use BLIP and CLIP to identify potential object categories from a predefined vocabulary in images. To enhance candidate set quality, we use CLIP with TagCLIP - a variant of CLIP tailored for multi-label classification. This improved pairing enables more accurate category predictions, resulting in a more reliable candidate set $\mathcal{C}$.

\subsection{Attention Extraction via Efficient Inversion} \label{sec:inversion}

\subsubsection{Preliminaries}

In text-conditioned diffusion models, such as Stable Diffusion \cite{sd}, generate an image by denoising a latent variable \( z_T \) under the condition of a text prompt $\mathcal{P}$. To enable this process, the network \( \epsilon_\theta \) is trained to predict the noise added during the forward diffusion process. The objective is to minimize the following loss function:
\begin{equation}
\min_{\theta} \mathbb{E}_{z_0, \epsilon\sim \mathcal{N}(0, 1), t \sim \text{Uniform}(1, T)}
\left\| \epsilon - \epsilon_{\theta}\bigl(z_t, t, \tau_{\theta}(\mathcal{P}) \bigl)\right\|_2^2.
\end{equation}
where $\tau_{\theta}(\mathcal{P})$ is the text embedding extracted from the prompt $\mathcal{P}$ via a text encoder $\tau_{\theta}$. In this formulation, the vector $z_t$ is a noisy sample, where noise is progressively added to the original data $z_0$ over T steps. During inference, the reverse denoising process is initiated from $z_T$, with the denoised results predicted sequentially by the pre-trained model \( \epsilon_\theta \). To accurately reconstruct a real image, we apply the deterministic DDIM sampling \cite{ddim} process. Specifically, the DDIM sampling process is as follows:
\begin{align}
z_{t-1} &= \sqrt{\alpha_{t-1}} \left[ \frac{z_t - \sqrt{1 - \alpha_t} \, \epsilon_{\theta}\bigl(z_t, t, \tau_{\theta}(\mathcal{P})\bigr)}{\sqrt{\alpha_t}} \right] \nonumber \\
&\quad + \sqrt{1 - \alpha_{t-1}} \epsilon_{\theta}\bigl(z_t, t, \tau_{\theta}(\mathcal{P})\bigr)
\end{align}

After recovering the latent \( z_0 \), the decoder \( D(\cdot) \) reconstructs the image \( I_{\text{gen}} = D(z_0) \). Unlike unconditional generation, text-conditioned inversion requires consistent alignment with the conditioning text, making reconstruction of the original image \( I_0 \) from \( z_T \) challenging and prone to inaccuracies. 

\subsubsection{Latent Inversion for Semantic Reconstruction}

\begin{figure}[]
	\centering
	\includegraphics[width=0.51\textwidth]{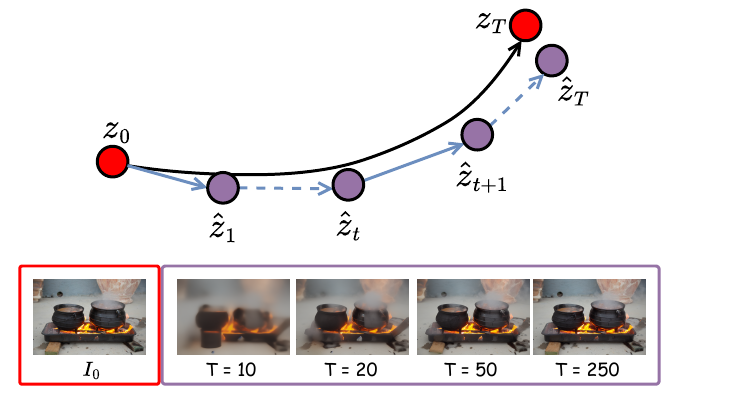}
	\caption{Effect of a number of DDIM inversion steps on reconstruction quality, where \( I_0 \) denotes the original input image. Increasing inversion steps improves reconstruction fidelity.}
	\label{fig:ddim}
\end{figure}

To enable segmentation using text-conditioned diffusion models, we first invert \( I_0 \) to recover the corresponding latent noise vector \( z_0 \) from the noisy latent \( z_T \). The inversion aims to find a pivotal noise vector from the forward noising process. This inversion process aims to map \( I_0 \) into the noise space of the diffusion model, enabling faithful reconstruction conditioned on the textual prompt $\mathcal{P}$. We adopt Deterministic Denoising Diffusion Implicit Models (DDIM) inversion~\cite{ddim,ddim_inversion} to reverse the forward diffusion process through the following iterative update:

\begin{align}
z_{t+1} &= \sqrt{\alpha_{t+1}} \left[
\frac{z_t - \sqrt{1-\alpha_t} \, \epsilon_{\theta}(z_t, t, \tau_{\theta}(\mathcal{P}))}{\sqrt{\alpha_t}} 
\right] \nonumber \\
&\quad + \sqrt{1-\alpha_{t+1}} \, \epsilon_{\theta}(z_t, t, \tau_{\theta}(\mathcal{P}))
\end{align}

In conventional diffusion sampling, classifier-free guidance is often applied to strengthen alignment between the generated image and the conditioning text by blending conditional and unconditional noise predictions:

\begin{align}
\epsilon_{\theta}(z_t, t, \tau_{\theta}(\cdot)) &= w \cdot \epsilon_{\theta}(z_t, t, \tau_{\theta}(\mathcal{P})) \nonumber \\
&\quad + (1 - w) \cdot \epsilon_{\theta}(z_t, t, \tau_{\theta}(\varnothing)),
\end{align}
where \( w > 1 \) controls the guidance strength. However, during inversion, using large guidance weights introduces semantic drift due to cumulative errors over multiple steps. Therefore, we fix \( w = 1 \) throughout the inversion process to maintain reconstruction stability.

Although DDIM inversion enables faithful reconstruction, it typically requires a large number of steps \( T \) to approximate the continuous reverse diffusion process accurately. As shown in Figure~\ref{fig:ddim}, reducing \( T \) results in significant reconstruction artifacts. This limitation stems from two factors: (1) The reverse diffusion process requires multiple small discrete steps to approximate the continuous probability flow ODE accurately. Large step sizes introduce instability and error accumulation. (2) The noise prediction network \( \epsilon_{\theta}(\cdot) \) is trained to operate on small step sizes; thus, larger inference steps reduce prediction accuracy.

Since DDIM inversion involves both forward and reverse passes, its computational complexity scales with 2T steps, posing a trade-off between reconstruction fidelity and inference time. To overcome this bottleneck, we employ 2-Rectified Flow (2-RF)~\cite{instaflow}, a distilled diffusion model built upon the Rectified Flow \cite{rf1,rf2} to enable faithful image inversion with minimal steps. 2-RF straightens the latent space ODE trajectories via a reflow procedure, enabling accurate inversion in a single run. Utilizing 2-RF allows accurate inversion with only two steps or (1+1)-step, yielding significant speed-up while preserving reconstruction quality.


\subsubsection{Attention Extraction for Segmentation}\label{sec:hard}

FA-Seg generates semantic segmentation masks by extracting cross-attention and self-attention maps during the Stable Diffusion denoising process (latent $z_1$ to $z_0$). The cross-attention maps $\mathcal{C}^{l,r} \in [0,1]^{r \times r \times d_{\mathcal{P}}}$ capture spatial correlations between text tokens and image regions. On the other hand, self-attention maps $\mathcal{S}^{l,r} \in [0,1]^{r \times r \times r \times r}$ preserve the spatial structure of the image by modeling pairwise similarities between image positions. These attention maps are computed across multiple resolutions $\mathcal{R} = \{8,16,32,64\}$, with each resolution having $L$ layers. The maps for each resolution are averaged over layers to obtain a final representation:

\begin{equation}
 \label{eq:5}
 \mathcal{C}^{r} = \frac{1}{L} \sum_{l=1}^L \text{Softmax} \left( \frac{Q^l_z K_{\tau_{\theta}(\mathcal{P})}^{\top}}{\sqrt{d}} \right).
\end{equation}

\begin{equation}
 \mathcal{S}^{r} = \frac{1}{L} \sum_{l=1}^L \text{Softmax} \left( \frac{Q^l_z K_z^{\top}}{\sqrt{d}} \right)
\end{equation}
where $Q_z$, $K_z$, and $K_{\tau_{\theta}(\mathcal{P})}$ represent the query and key matrices derived from linear projections. By using a class prompt \( \mathcal{P}' \) we control the cross-attention for the candidate classes rather than relying solely on the text prompt \( \mathcal{P} \). While \( \mathcal{P} \) continues to guide image reconstruction via DDIM inversion, we propose using \( \mathcal{P}' \) to generates segmentation masks for all candidate classes and offers greater flexibility in specifying candidate classes compared to the text prompt \( \mathcal{P} \). This dual-prompt enables the diffusion model to simultaneously reconstruct the image while injecting class-awareness into attention maps. In this setting, cross-attention maps \( \mathcal{C}^{l} \) in Equation \ref{eq:5} are computed using \( K_{\tau_{\theta}(\mathcal{P'})} \) instead of \( K_{\tau_{\theta}(\mathcal{P})} \). This approach enables the generation of cross-attention maps for all candidate classes within a single inference run.

\subsection{Hierarchical Attention Refinement Method}\label{sec:refinement}

\begin{figure}[]
    \centering
    \begin{subfigure}[b]{0.48\textwidth}
        \centering
        \includegraphics[width=\textwidth]{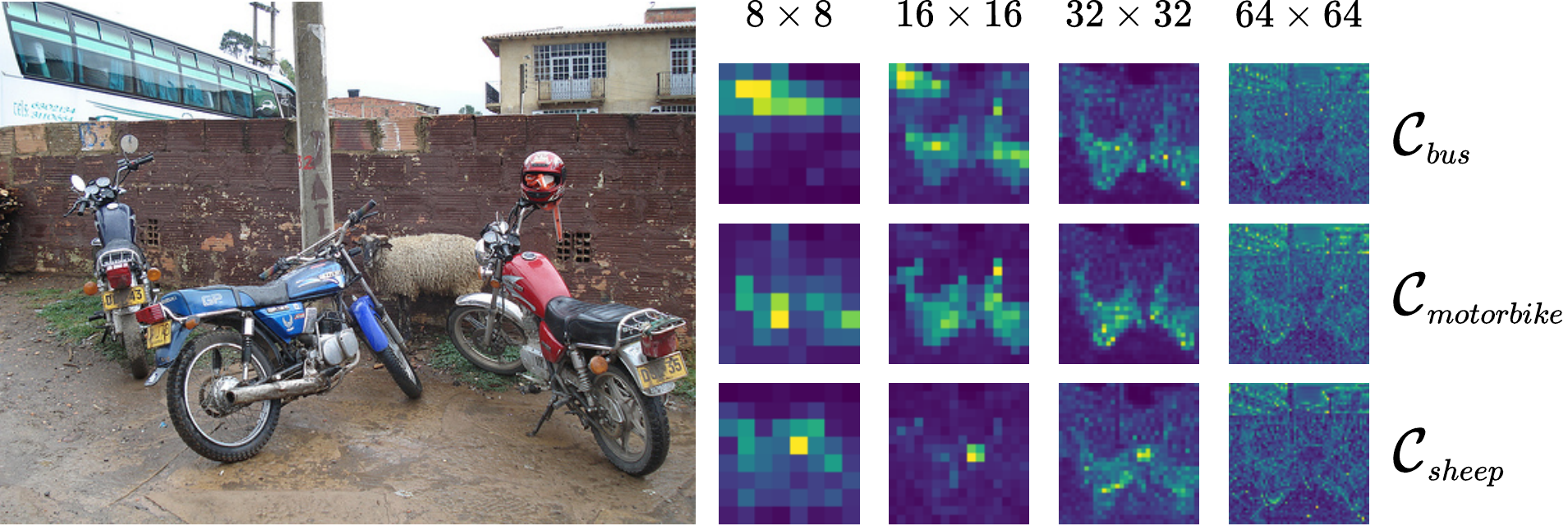}
        \caption{Visualization of cross-attention maps for the classes \texttt{bus}, \texttt{motorbike} and \texttt{sheep} at different resolutions.}
        \label{fig:cross}
    \end{subfigure}
    
    \begin{subfigure}[b]{0.48\textwidth}
        \centering
        \includegraphics[width=\textwidth]{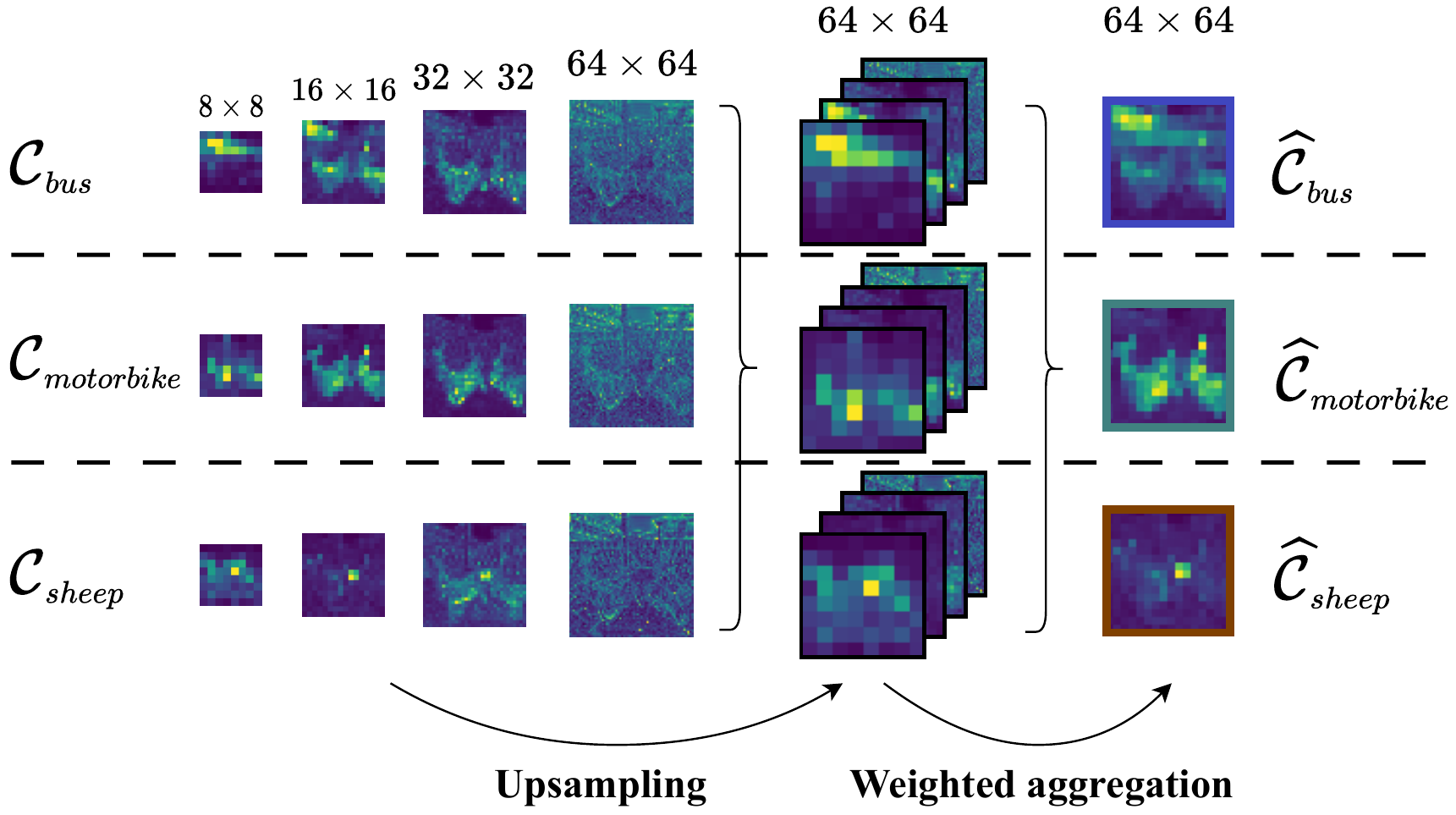}
	\caption{The process of aggregating cross-attention maps for three classes: \textit{bus}, \textit{motorbike}, and \textit{sheep} across different resolutions. Initially, all cross-attention maps are upsampled to a common resolution of $64 \times 64$. Following this, the upsampled maps are aggregated using a weighted scheme in order to produce the final fused attention map for each class.} 
	\label{fig:cross_fusion}%
    \end{subfigure}
    
    \caption{Cross-attention visualization.}
\end{figure}

In semantic segmentation tasks, attention maps play a crucial role in localizing and identifying different regions within an image. However, attention maps at a fixed resolution provide valuable localization information but may fail to capture the semantic diversity present across multiple spatial scales. As illustrated in Figure~\ref{fig:cross}, cross-attention maps at various resolutions offer complementary insights: $\mathcal{C}^{8}$ provides coarse maps but accurate localization, while $\mathcal{C}^{16}$ captures better details of object position and shape. At higher resolutions (e.g., $\mathcal{C}^{32}$ and $\mathcal{C}^{64}$), attention maps may tend to degrade in quality and become increasingly complex to make sense of. These observations align with previous studies \cite{Pplus}, showing that higher-resolution layers often focus on superficial features (such as color or style), while lower-resolution layers encode object semantics. To address this limitation, we are using a hierarchical cross-attention fusion mechanism that emphasizes semantically cross-attention maps (e.g., $\mathcal{C}^{16}$) while down-weighting noisy maps (e.g., $\mathcal{C}^{32}$ or $\mathcal{C}^{64}$). The fused cross-attention map $\widehat{\mathcal{C}}_c$ is computed as a weighted sum of the resolution-specific maps, each upsampled to a size of $64 \times 64$:

\begin{equation}
\widehat{\mathcal{C}_c} = \sum_{r \in \mathcal{R}} w^r_{\text{cross}} \times \text{Upsample}\left(\mathcal{C}^{r}_c, (64, 64)\right)
\end{equation}

\begin{figure*}[]
	\centering 
	\includegraphics[width=0.9\textwidth]{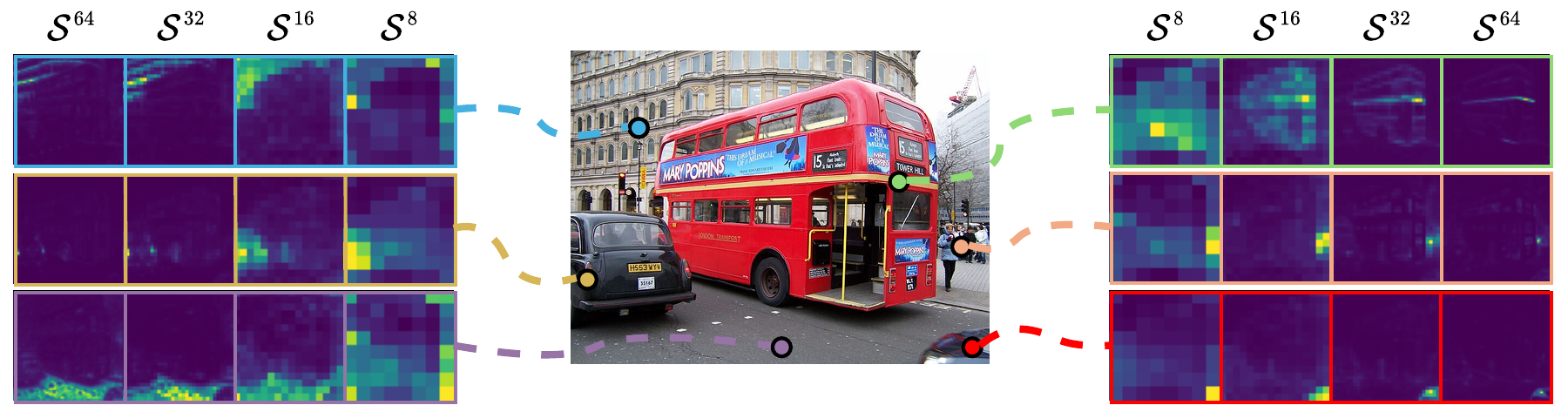}	
	\caption{\textbf{Self-attention maps at different resolutions within the SD model.} We visualize the self-attention maps of six query points, each marked with a distinct color. The results show that each attention map focuses on a localized region of the input image. Specifically, self-attention maps with lower resolutions (8$\times$8, 16$\times$16) represent a larger receptive field, allowing for a more general representation of object groups. In contrast, self-attention maps with larger resolutions (32$\times$32, 64$\times$64) can identify smaller objects and finer details within the objects.} 
	\label{fig:self}%
\end{figure*}

The upsampling and weighted aggregation of cross-attention maps at multiple resolutions are demonstrated in Figure~\ref{fig:cross_fusion} for three candidate classes: \texttt{bus}, \texttt{motorbike}, and \texttt{sheep}. Although the fused cross-attention map $\widehat{\mathcal{C}}_c$ accurately highlights the spatial locations of the target class, it remains coarse and lacks detailed structure. To refine $\widehat{\mathcal{C}}_c$, we incorporate self-attention maps $\mathcal{S}^r$ from multiple resolutions, which provide complementary structural information at various spatial scales. This combination of self-attention and cross-attention maps across multiple resolutions is the core of our \textbf{H}ierarchical \textbf{A}ttention \textbf{R}efinement Metho\textbf{d} (\textbf{HARD}).

\begin{figure}[]
	\centering 
	\includegraphics[width=0.44\textwidth]{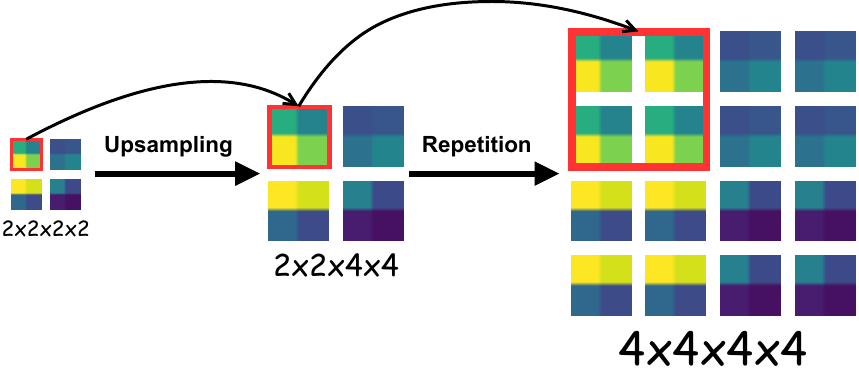}	
	\caption{An example of the transformation process from a 4D tensor of shape $2 \times 2 \times 2 \times 2$ to a tensor of shape $4 \times 4 \times 4 \times 4$. This transformation is performed in two steps: upsampling followed by repetition.} 
	\label{fig:up_and_repeat}%
\end{figure}

Each self-attention tensor $\mathcal{S}^r \in [0,1]^{r \times r \times r \times r}$ captures pairwise affinities among all spatial locations at resolution $r$. In a query-wise perspective, each spatial query position $(i, j)$ corresponds to a dedicated 2D attention map $\mathcal{S}^r_{[i,j,:,:]} \in [0,1]^{r \times r}$, which captures the semantic context relevant to that specific query. Figure~\ref{fig:self} illustrates query-wise self-attention maps across different resolutions and spatial locations. The resolution of each query-wise attention map defines its receptive field in relation to the original image. To incorporate self-attention information into the segmentation process, we refine the fused cross-attention map $\widehat{\mathcal{C}_c}$ by applying affinity weighting derived from query-wise self-attention maps. Since $\widehat{\mathcal{C}_c} \in \mathbb{R}^{64 \times 64}$ is defined at a fixed spatial resolution, we transform the original self-attention tensors $\mathcal{S}^r$ into $\mathcal{S}^{r \rightarrow 64} \in \mathbb{R}^{64 \times 64 \times 64 \times 64}$ through upsampling and repetition. This transformation is illustrated in Figure~\ref{fig:up_and_repeat}, with an example from shape $2 \times 2 \times 2 \times 2$ to $4 \times 4 \times 4 \times 4$. While upsampling the last two dimensions is straightforward, repeating the first two dimensions relies on the empirical observation: 2D attention maps at neighboring positions often correspond to the same object group (e.g., \(\mathcal{S}^r_{[i,j,:,:]}\) and \(\mathcal{S}^r_{[i+1,j,:,:]}\), or \(\mathcal{S}^r_{[i,j,:,:]}\) and \(\mathcal{S}^r_{[i,j+1,:,:]}\)) thus allowing spatial continuity to be preserved.

This transformation produces spatially aligned 4D self-attention tensors \(\mathcal{S}^{r \rightarrow 64}\), which are compatible for multiplicative refinement of the class-specific cross-attention map \(\widehat{\mathcal{C}}_c\). To incorporate self-attention affinities at multiple resolutions \(r \in \mathcal{R}\), each \(\mathcal{S}^{r \rightarrow 64}\) is matricized into an affinity matrix, and \(\widehat{\mathcal{C}}_c\) is vectorized to match this dimension. These maps are then normalized and combined via a weighted sum over all resolutions to yield the final refined segmentation mask:

\begin{equation}
\widehat{\mathcal{A}}_c = \sum_{r \in \mathcal{R}} w^r_{self} \times \text{norm}\Bigl(\text{matricize}(\mathcal{S}^{r \rightarrow 64}) \cdot \text{vectorize}(\widehat{\mathcal{C}}_c)\Bigl)
\end{equation}

By calculating $\widehat{\mathcal{A}}$ for the entire candidate class list $C$, we obtain the set of segmentation maps $\mathbf{A} = \{\widehat{\mathcal{A}}_c \mid c \in C\}$. Then, by applying the Test-Time Flipping (TTF) technique, we obtain two corresponding segmentation map lists: $\mathbf{A}^{\text{origin}}$ from the original image and $\mathbf{A}^{\text{flipped}}$ from the horizontally flipped image. We align the spatial information of the attention maps from the flipped image using $\mathcal{F}lip$ to match those from the original image. Then, we compute the average of each pair of corresponding attention maps from both lists to improve spatial consistency and prediction accuracy.

\[
\widehat{\mathcal{A}}_c^* = \frac{1}{2} \left[ \widehat{\mathcal{A}}_c^{\text{origin}} + \mathcal{F}lip\Bigl(\widehat{\mathcal{A}}_c^{\text{flipped}}\Bigl) \right]
\]

To produce the final segmentation mask, the score maps in the list $A^*$ obtained after TTF are subjected to the $\arg\max$ operation along the class dimension to assign a class label to each pixel. To identify background regions, a threshold $\alpha \in [0, 1]$ is applied: pixels with maximum scores lower than $\alpha$ are assigned to the background class, while those exceeding the threshold are assigned to the corresponding class label.

\section{Experiments}

\begin{table*}[h]
\centering
\caption{Comparison with existing methods. Models in the first three rows are finetuned on target datasets while the rest
approaches do not require mask annotations. \textbf{Bold} fonts refer to the best results and \underline{underline} fonts refer to the second best. Note that ``\xmark'' refers to the training-free method, and ``--'' represents unprovided data.}
\label{tab:main_results}
\begin{tabular}{lccccc}
\toprule
\multirow{2}{*}{\textbf{Methods}} & \multirow{2}{*}{\textbf{Training datasets}} & \multicolumn{4}{c}{\textbf{mIOU}}                                                                                                                                                                                     \\ \cmidrule{3-6} 
                                  &                                                                                       & \textbf{PASCAL VOC} & \textbf{PASCAL Context} & \textbf{COCO Object} & \textbf{Avg.} \\ 
\midrule
\multicolumn{6}{l}{\textit{Contrastive learning based}}  \\
\midrule
MaskCLIP \cite{maskclip} \textit{\scalebox{0.7}{(ECCV '22)}} & \xmark & 29.3 & 21.1 & 15.5 & 22.0 \\
GroupViT \cite{GroupViT} \textit{\scalebox{0.7}{(CVPR '22)}}               & CC12M                                                                                 & 50.4                & 18.7                    & 27.5                 & 32.2          \\
ReCo \cite{reco} \textit{\scalebox{0.7}{(NeurIPS '22)}}                & \xmark                                                                                    & 25.1                & 19.9                    & 15.7                 & 20.2          \\
TCL \cite{tcl} \textit{\scalebox{0.7}{(CVPR '23)}}                    & CC3M+CC12M                                                                            & 51.2                & 24.3                    & 30.4                 & 35.3          \\
ZeroSeg \cite{ZeroSeg} \textit{\scalebox{0.7}{(ICCV '23)}}                & IN-1K                                                                                 & 40.8                & 20.4                    & 20.2                 & 27.1          \\
ViewCo \cite{viewco} \textit{\scalebox{0.7}{(ICLR '23)}}                 & CC12M+YFCC                                                                            & 52.4                & 23.0                    & 23.5                 & 33.0          \\
SegCLIP \cite{SegCLIP} \textit{\scalebox{0.7}{(ICML '23)}}                & CC3M+COCO                                                                             & 52.6                & 24.7                    & 26.5                 & 34.6          \\
OVSegmentor \cite{ovsegmentor} \textit{\scalebox{0.7}{(CVPR '23)}}            & CC4M                                                                                  & 53.8                & 20.4                    & 25.1                 & 33.1          \\
SimSeg \cite{simseg} \textit{\scalebox{0.7}{(CVPR '23)}}                 & CC3M+CC12M                                                                            & 57.4                & 26.2                    & 29.7                 & 37.8          \\
SAM-CLIP \cite{samclip} \textit{\scalebox{0.7}{(CVPRW '24)}}                 & Merged-41M    &    60.6    &     29.2      &     31.5                 & 40.4          \\
CLIP-DIY \cite{CLIP-DIY} \textit{\scalebox{0.7}{(WACV '24)}} & \xmark & 59.9 & 19.7 & 31.0 & 36.9 \\
SCLIP \cite{sclip} \textit{\scalebox{0.7}{(ECCV '24)}}                  & \xmark                                                                                   & 59.1                & 30.4        & 30.5                 & 40.0          \\ 
ClearCLIP \cite{clearclip} \textit{\scalebox{0.7}{(ECCV '24)}} & \xmark & 51.8 & \textbf{32.6} & 33.0 & 39.1 \\
CLIP Surgery \cite{CLIPSurgery} \textit{\scalebox{0.7}{(Pattern Recognition '25)}} & \xmark & -- & 29.3 & -- & -- \\
\midrule
\multicolumn{6}{l}{\textit{Diffusion models based}}                                                                                                                \\ \midrule
Diffusion Baseline \cite{sd} \textit{\scalebox{0.7}{(CVPR '22)}}     & \xmark                                                                                    & 59.6                & 25.0                    & 34.5                 & 39.7          \\
FreeSeg-Diff \cite{FreeSegDiff} \textit{\scalebox{0.7}{(arXiv '24)}}          & \xmark                                                                                    & 53.3                & --                      & 31.0                 & --            \\
DiffCut  \cite{diffcut} \textit{\scalebox{0.7}{(NeurIPS '2024)}}                & \xmark & 63.0 & 24.6& 36.0 & 41.2 \\      
Semantic DiffSeg \cite{diffseg_un} \textit{\scalebox{0.7}{(CVPR '24)}}                & \xmark                                                                                    & 39.4                & 16.7                    & 19.1                 & 25.1          \\
DiffSegmenter \cite{diffsegmenter} \textit{\scalebox{0.7}{(TIP '25)}}           & \xmark                                                                                   & 60.1                & 27.5                    & \textbf{37.9}        & 41.8          \\
InvSeg \cite{InvSeg} \textit{\scalebox{0.7}{(AAAI '25)}}                  & \xmark                                                                                   & \underline{63.4}                & 27.8                    & 36.0                 & \underline{42.4} \\ \midrule
\textbf{FA-Seg (Ours)}                     & \xmark                                                                          & \textbf{64.0}       & \underline{31.3} & \underline{36.2} & \textbf{43.8} \\ \bottomrule
\end{tabular}
\end{table*}

\subsection{Datasets and Metric}

We evaluate the proposed FA-Seg on standard benchmark datasets for the Open-Vocabulary Semantic Segmentation (OVSS) task: PASCAL VOC \cite{voc}, PASCAL Context \cite{Context}, and MS COCO Object \cite{coco}, comprising 20, 59, and 80 object categories, respectively (excluding the background class). As FA-Seg is a training-free method, evaluation is conducted directly on the validation sets of each dataset, which include 1,449 images from PASCAL VOC, 5,004 images from PASCAL Context, and 5,000 images from MS COCO Object. Following prior training-free diffusion-based OVSS studies \cite{diffsegmenter,InvSeg,diffcut}, we adopt these datasets to ensure a fair and consistent comparison with existing approaches. To assess segmentation performance, we employ the mean Intersection-over-Union (mIoU) metric, which is a widely used standard for evaluating semantic segmentation models that measure the overlap between predicted and ground-truth regions across classes.

\subsection{Implementation Details}
We employ 2-Rectified Flow \cite{instaflow}, a distilled variant of Stable Diffusion 1.5 (SD 1.5), as the primary model for attention map extraction. All inferences of SD are performed at an image resolution of 512$\times$512 pixels. In the multi-resolution attention fusion, we assign distinct weights to cross-attention and self-attention maps. Specifically, for cross-attention maps, we use the weight vector $w^r_{\text{cross}} \in {0.15, 0.7, 0.15, 0}$ corresponding to resolution levels $r \in {8, 16, 32, 64}$, while for self-attention maps, we assign weights of $w^r_{\text{self}} \in {0.1, 0.1, 0.5, 0.3}$. These weights were selected based on both qualitative observations of the attention maps (Section \ref{sec:hard}) and quantitative results (Table~\ref{tab:ab_diff}). Attention maps contributing more strongly to semantic representation are assigned higher weights, whereas those from less informative resolutions are down-weighted or excluded from the aggregation. It is important to note that these weights were not obtained through exhaustive hyperparameter search. Although they may not represent the globally optimal configuration, the consistently improved results validate the effectiveness of the proposed attention-weighted aggregation strategy. All experiments are implemented in PyTorch with a batch size of 2 and conducted on a system equipped with an NVIDIA GeForce RTX 4090 GPU (32GB RAM)


\subsection{Main Results}
\subsubsection{Quantitative results}

Table~\ref{tab:main_results} summarizes the performance of our proposed FA-Seg method across three widely-used open-vocabulary semantic segmentation benchmarks: PASCAL VOC 2012, PASCAL Context, and COCO Object. FA-Seg achieves mIoU scores of 64.0\% on PASCAL VOC, 31.3\% on PASCAL Context, and 36.2\% on COCO Object, leading to the highest average mIoU among training-free diffusion-based approaches across all three datasets. While results from contrastive vision-language models are also reported for comprehensive comparison, our primary focus centers on a direct and fair comparison with recent training-free diffusion-based methods, including Diffusion Baseline~\cite{sd}, FreeSeg-Diff~\cite{FreeSegDiff}, DiffCut~\cite{diffcut}, DiffSeg~\cite{diffseg_un}, DiffSegmenter~\cite{diffsegmenter}, and InvSeg~\cite{InvSeg}. The performance of Diffusion Baseline is referenced from the InvSeg study~\cite{InvSeg}, while the results of Semantic DiffSeg, an adaptation of DiffSeg \cite{diffseg_un} for open-vocabulary settings, are taken from its corresponding publication~\cite{diffsegmenter}. FA-Seg establishes state-of-the-art accuracy on PASCAL VOC 2012 and delivers the highest overall average performance across all datasets. On PASCAL Context and COCO Object, FA-Seg ranks second, with results only marginally trailing behind the best-performing methods on each dataset.  These strong results stem from our proposed methods: the \textit{dual-prompt mechanism} improves class-aware attention control, the \textit{HARD} module enhances object boundary accuracy through multi-resolution attention fusion, and the \textit{TTF} strategy further improves spatial consistency during inference.

\begin{table*}[!h]
\centering
\caption{Comparison of cost of existing diffusion-based segmentation methods.}
\label{tab:time_mem}
\begin{tabular}{lccccc}
\toprule
              & \textbf{GPU device} & \textbf{Optimization step} & \textbf{Memory} & \textbf{Inference time} & \textbf{Avg. mIoU} \\ \midrule
InvSeg \cite{InvSeg}        & H100                & \checkmark                 & 32.4G           & 7.9s                    & 42.4               \\
DiffSegmentor \cite{diffsegmenter} & RTX 4090            & \xmark                     & 16.1G           & M$\times$0.44s          & 41.8               \\
\textbf{FA-Seg (Ours)} & RTX 4090            & \xmark                     & 13.4G           & 0.36s                   & 43.8               \\ \bottomrule
\end{tabular}%
\end{table*}

\begin{table*}[]
\centering
\caption{Ablation study on the contributions of each component in FA-Seg, including self-attention refinement, cross-attention weighted aggregation, self-attention weighted aggregation, class prompt design, and test-time flipping. Removing any individual component leads to decreased performance.}
\label{tab:ab_component}
\begin{tabular}{lcccc}
\toprule
\multirow{2}{*}{\textbf{FA-Seg}}                 & \textbf{PASCAL VOC} & \textbf{PASCAL Context} & \textbf{COCO Object} & $\downarrow\Delta$ \\ \cmidrule{2-5} 
                                                  & 64.0                & 31.3                    & 36.2                 & -        \\ \midrule
\textit{w/o self attention refinement}            & 39.6                & 25.3                     & 21.3                  & 15.1      \\
\textit{w/o cross attention weighted aggregation} & 61.2                & 30.2                    & 35.0                 & 1.7      \\
\textit{w/o self attention weighted aggregation}  & 63.0                & 30.9                    & 35.8                 & 0.6      \\
\textit{w/o dual-prompt mechanism}                         & 62.5                & 31.0                    & 36.2                 & 0.6      \\
\textit{w/o test-time flipped}                    & 63.6                & 31.0                    & 36.0                 & 0.3      \\ \bottomrule
\end{tabular}%
\end{table*}


Table \ref{tab:time_mem} presents a quantitative evaluation of the operational efficiency (exclude candidate class extraction) among existing state-of-the-art diffusion-based segmentation methods, including DiffSegmentor, InvSeg, and our FA-Seg . In terms of memory consumption, FA-Seg demonstrates superior efficiency, utilizing only 13.4GB, significantly less than DiffSegmentor's 16.1GB and InvSeg's 32.4GB . Notably, regarding inference time, FA-Seg achieves superior performance with just 0.36 seconds on an RTX 4090 GPU. This figure represents a substantial improvement over DiffSegmentor, whose inference time depends on the number of candidate classes \textit{M}, and is particularly impressive when compared to InvSeg, which requires up to 7.9 seconds per image on an NVIDIA H100 GPU due to its complex prompt embedding optimization procedure. This efficiency stems from FA-Seg's ability to perform fast semantic reconstruction for all candidate classes in a single inference run. Furthermore, to maximize throughput and scalability, all evaluations are conducted using half-precision inference. While this introduces a marginal trade-off in numerical precision, it results in considerable gains in speed and memory usage. These results demonstrate that FA-Seg offers high inference efficiency while significantly reducing the necessary computational resources.

In summary, FA-Seg not only achieves state-of-the-art segmentation performance among training-free diffusion-based methods but also demonstrates superior inference efficiency and deployment simplicity, making it an efficient and scalable solution for real-world open-vocabulary semantic segmentation.

\begin{figure*}
	\centering 
	\includegraphics[width=0.952\textwidth]{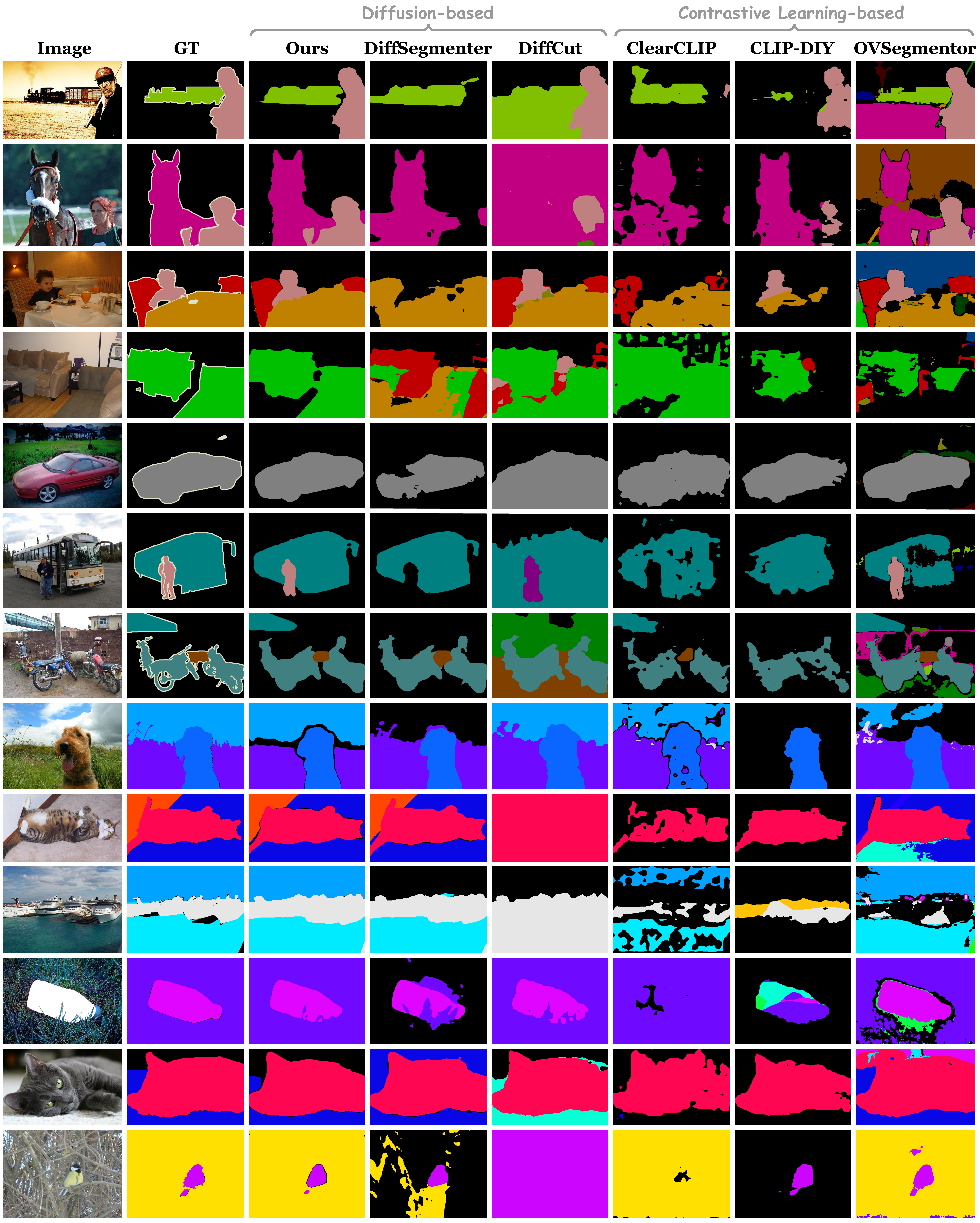}	
	\caption{Open vocabulary semantic segmentation results of FA-Seg and DiffSegmenter \cite{diffsegmenter}, DiffCut \cite{diffcut}, ClearCLIP \cite{clearclip}, CLIP-DIY \cite{CLIP-DIY}, OVSegmentor \cite{ovsegmentor}} 
	\label{fig:qualitative_results}%
\end{figure*}

\subsubsection{Qualitative results}

Figure~\ref{fig:qualitative_results} provides a qualitative comparison of FA-Seg against several methods, including DiffSegmenter, DiffCut, ClearCLIP, CLIP-DIY, and OVSegmentor, on two benchmark datasets: PASCAL VOC 2012 (top 7 rows) and PASCAL Context (bottom 6 rows). To ensure fairness, the PAMR post-processing step~\cite{pamr} was removed when showing DiffCut results. The results highlight several key observations. DiffSegmenter performs well in capturing object shapes on PASCAL VOC but struggles with contextual classes (e.g., \textit{tree}, \textit{glass}) in PASCAL Context. Additionally, its performance is highly sensitive to the quality of the candidate class list. DiffCut produces unstable segmentation masks, attributed to its reliance on an unsupervised grouping strategy and requiring FC-CLIP for semantic label association, making it suboptimal for open-vocabulary segmentation tasks. Among CLIP-based methods, CLIP-DIY often suffers from false positives and false negatives, while ClearCLIP partially mitigates this issue but still shows failure cases-particularly missing detections for classes like \textit{person}. This indicates limited spatial localization capacity in these models. OVSegmentor exhibits strong geometric segmentation but lacks effective mechanisms to filter out irrelevant or negative classes, leading to noisy outputs and reduced precision.

In contrast, FA-Seg consistently delivers high-quality segmentation masks, demonstrating both accurate geometric localization and precise category alignment, even in challenging scenarios. These results substantiate FA-Seg’s advantage in open-vocabulary settings, offering a strong balance between semantic accuracy and spatial precision.

\subsection{Ablation study}

\subsubsection{Effects of different components}

Table ~\ref{tab:ab_component} presents an ablation study that evaluates the contribution of components in FA-Seg. Removing the cross-attention and self-attention weighted aggregation modules is equivalent to using only a single-resolution attention map with the best performance (as shown in Table ~\ref{tab:ab_diff}), without applying multi-resolution aggregation-specifically, $16 \times 16$ for cross-attention and $32 \times 32 \times 32 \times 32$ for self-attention. In contrast, removing the class prompt $\mathcal{P}'$ means that cross-attention maps are computed solely based on the text prompt $\mathcal{P}$, without explicitly guiding the model to focus on candidate classes. In the original configuration, the class candidate list is appended to the image caption to form the complete prompt $\mathcal{P}$, allowing the model to utilize the appended candidate list to generate more targeted cross-attention maps. Among all components, self-attention refinement proves to be the most critical, as its removal leads to the most significant average performance drop (-15.1\%), underscoring its importance in enhancing the spatial precision of class-aware attention. Both cross-attention and self-attention weighted aggregation also play important roles, as their removal results in noticeable performance degradation. On the other hand, class prompt design and test-time flipping (TTF) have relatively more minor effects-0.6\% and 0.3\%, respectively-yet consistently contribute to overall segmentation quality.

These findings validate the overall design of FA-Seg, where each component serves a complementary role in improving segmentation performance while maintaining computational efficiency.

\subsubsection{Different background threshold}

\begin{figure}[]
\centering

\begin{minipage}{0.50\textwidth}
\centering
\captionof{table}{mIoU results from threshold 0.4 to 0.65 for COCO, Context, and VOC datasets}
\label{tab:thread}
\begin{tabular}{lcccccc}
    \toprule
    \textbf{Dataset}  & 0.4 & 0.45 & 0.5 & 0.55 & 0.6 & 0.65 \\
    \midrule
    PASCAL VOC     & 59.9 & 62.4 & 63.8 & \textbf{64.0} & 62.3 & 58.6 \\
    PASCAL Context & 30.8 & 31.2 & \textbf{31.3} & 30.9 & 29.8 & 28.1 \\
    COCO Object    & 32.5 & 34.2 & 35.5 & \textbf{36.2} & \textbf{36.2} & 35.3 \\
    \bottomrule
\end{tabular}
\end{minipage}
\begin{minipage}{0.48\textwidth}
\centering
\includegraphics[width=\textwidth]{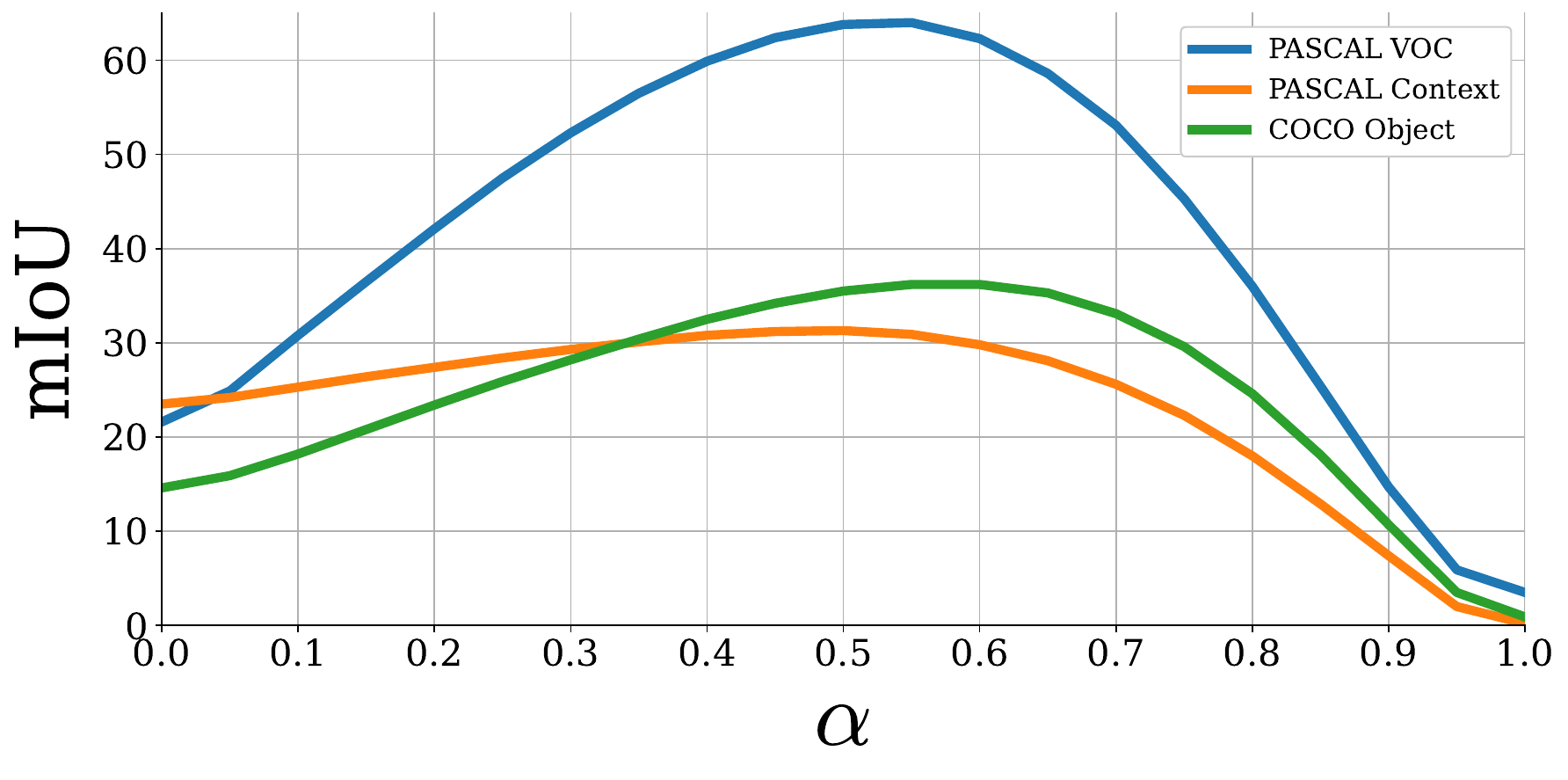}
\caption{Comparison of mIoU across different threshold values for three datasets.}
\label{fig:thread}
\end{minipage}
\end{figure}

Since the background class is not inferred directly from the model, we determine background regions by applying a threshold to the predicted score maps. Specifically, pixels with score values below the selected threshold are assigned as background. To examine the impact of this threshold on segmentation performance, we visualize the mIoU scores corresponding to thresholds ranging from 0 to 1 with a step size of 0.05, as shown in Figure~\ref{fig:thread}. In addition, to enable a more focused analysis, we report detailed mIoU values for thresholds between 0.4 and 0.65-identified as the range yielding the best results-in Table~\ref{tab:thread}. Thresholds in the range of 0.5 to 0.6 generally achieve optimal segmentation performance depending on the dataset. The performance tends to increase with higher thresholds up to a certain point, after which it gradually declines, suggesting that the model is stable under small perturbations of this parameter.

\subsubsection{Different class candidate generation strategies.}

In contrast to Contrastive Language-Image Pre-Training, open-vocabulary segmentation methods that rely on diffusion models require a candidate class list to be provided at inference rather than segmenting over a predefined set of classes. Table~\ref{tab:class} reports the segmentation performance under different candidate list generation strategies across three benchmark datasets. In addition, we also report results using the ground-truth class list as a class candidate list. The results indicate that combined methods such as CLIP+TagCLIP and BLIP+CLIP outperform single-model baselines, including BLIP, CLIP, and TagCLIP. While our proposed method, TagCLIP+BLIP, does not consistently achieve the highest performance on every individual dataset, it demonstrates the most stable performance across all three benchmarks, suggesting its robustness and generalizability. Furthermore, the results obtained with the ground-truth class list highlight the potential for further improvement in segmentation accuracy when more precise candidate lists are generated.

\begin{table}[!t]
\caption{Ablating different class candidate generation strategies.}
\label{tab:class}
\resizebox{1\columnwidth}{!}{%
\begin{tabular}{lccc}
\toprule
             & \textbf{\small{PASCAL VOC}}    & \textbf{\small{VOC Context}}   & \textbf{\small{COCO Object}}   \\ \midrule
BIP          & 58.3          & 26.4          & 34.3          \\
CLIP         & 63.0          & 23.9          & 34.8          \\
TagCLIP      & \textbf{65.0} & \underline{29.3}          & 34.8          \\
CLIP+BIP     & 62.8          & 28.8 & \textbf{36.7} \\
TagCLIP+BIP  & \underline{64.0}          & \textbf{31.3} & \underline{36.2}          \\ \midrule
\textbf{\underline{Ground Truth}} & \textbf{\underline{69.7}}          & \textbf{\underline{45.1}}          & \textbf{\underline{38.8}}          \\ \bottomrule
\end{tabular}%
}
\end{table}

\subsubsection{Effect of different attention scales}

Table \ref{tab:ab_diff} presents a comprehensive study on the effect of different attention scales and fusion strategies in FA-Seg. The results reveal several key findings. First, single-resolution attention maps, particularly those from the 16×16 resolution level, consistently yield higher segmentation performance across all datasets compared to other individual scales, suggesting that this resolution offers a favorable balance between semantic context and spatial detail. Second, when combining attention maps across multiple resolutions, both mean and weighted aggregation strategies improve overall performance, highlighting the complementary nature of multi-scale attention. Notably, weighted fusion outperforms mean fusion, as it enables the model to emphasize semantically rich and informative layers (e.g., 16×16 and 32×32) while down-weighting noisier or less discriminative ones (e.g., 64×64). The best overall performance-with an average mIoU of 64.0 / 36.2 / 31.3 on PASCAL VOC, Context, and COCO, respectively-is achieved using the weighted fusion of both cross- and self-attention maps, as indicated by the red-highlighted entry. Finally, attention maps at very low (8×8) or very high (64×64) resolutions individually underperform, reaffirming that mid-level resolutions are the most effective for capturing meaningful object-level features. These findings validate the design choice in FA-Seg to apply multi-resolution attention fusion, with empirically optimized weights, to maximize segmentation accuracy across diverse datasets.

\begin{table*}[]
\centering
\caption{Study on different attention scales and fusion strategies. The $^\textbf{1}$, $^\textbf{2}$, and $^\textbf{3}$ entries indicate the best, second-best, and third-best results, respectively.}
\label{tab:ab_diff}
\resizebox{2.07\columnwidth}{!}{%
\begin{tabular}{ccccccc}
\toprule
\multirow{2}{*}{\textbf{\begin{tabular}[c]{@{}c@{}}Cross\\ attention\end{tabular}}} & \multicolumn{6}{c}{\textbf{Self-attention}}                                              \\ \cmidrule{2-7} 
                                                                                    & \textbf{8} & \textbf{16} & \textbf{32} & \textbf{64} & \textbf{Mean} & \textbf{Weighted} \\ \midrule
\textbf{8}                                                                          & 50.5 / 26.1 / 23.9  & 56.1 / 30.9 / 27.1   & 54.4 / 30.6 / 26.5   & 53.4 / 29.3 / 26.0   & 57.3 / 31.2 / 27.5     & 55.9 / 30.9 / 27.0         \\
\textbf{16}                                                                         & 52.1 / 27.1 / 24.8  & 59.3 / 34.3 / 29.4   & 58.7 / 34.0 / 29.4   & 58.7 / 33.0 / 29.1   & 62.4 / 35.5 / 30.5     & 61.2 / 35.0 / 30.2         \\
\textbf{32}                                                                         & 46.0 / 20.4 / 15.2  & 46.0 / 21.8 / 16.4   & 34.9 / 17.8 / 13.2   & 37.1 / 18.3 / 13.7   & 46.0 / 21.3 / 15.9     & 40.0 / 19.4 / 14.4         \\
\textbf{64}                                                                         & 17.3 / \textcolor{white}{0}9.6 / \textcolor{white}{0}8.1   & 16.7 / \textcolor{white}{0}9.2 / \textcolor{white}{0}7.9    & 15.0 / \textcolor{white}{0}7.7 / \textcolor{white}{0}7.4    & 13.1 / \textcolor{white}{0}6.6 / \textcolor{white}{0}6.4    & 16.4 / \textcolor{white}{0}8.5 / \textcolor{white}{0}7.6      & 15.2 / \textcolor{white}{0}7.7 / \textcolor{white}{0}7.3          \\
\textbf{Mean}                                                                       & 51.9 / 27.1 / 24.6  & 60.0 / 33.5 / 29.1   & 61.0 / 34.5 / 29.5   & 59.8 / 33.1 / 28.9   & 62.4 / 34.5 / 29.8     & 62.0 / 34.7 / 29.9         \\
\textbf{Weighted}                                                                   & 52.1 / 27.5 / 25.3  & 61.0 / 34.8 / 30.1   & 63.0$^\textbf{3}$ / 35.8$^\textbf{3}$ / 30.9$^\textbf{3}$   & 61.9 / 34.7 / 30.3   & 63.8$^\textbf{2}$ / 36.0$^\textbf{2}$ / 31.1$^\textbf{2}$     & 64.0$^\textbf{1}$ / 36.2$^\textbf{1}$ / 31.3$^\textbf{1}$         \\ \bottomrule
\end{tabular}%
}
\end{table*}

\subsubsection{Different text prompt $\mathcal{P}$.}

To assess the impact of image description quality on FA-Seg's segmentation performance, we experiment with several text prompt generation strategies for constructing the reconstruction prompt $\mathcal{P}$. In addition to a \textit{simple text prompt}-“A photo of \{\textit{class candidates}\}”-which does not rely on any image captioning model, we evaluate a variety of automatic captioning methods. These include BLIP \cite{blip1}, used as the default captioning model in our framework, as well as more recent models such as BLIP-2 \cite{blip2}, GIT \cite{GIT}, and Gemini Flash 2.0. For Gemini, we consider two prompting strategies: Gemini$^{(1)}$ uses the instruction “Describe this image in one sentence” to generate a full description, whereas Gemini$^{(2)}$ uses “Describe the image in a short and generic sentence, avoid specific names or proper nouns” to generate more abstract, less detailed captions. As shown in Table~\ref{tab:prompt}, the model exhibits relatively stable performance across different captioning methods, indicating that FA-Seg is not overly sensitive to variations in prompt content. Interestingly, we observe that overly detailed prompts (Gemini$^{(1)}$) slightly degrade performance, whereas more simple prompts (Gemini$^{(2)}$) yield more better results. Additionally, the \textit{simple text prompt} also yields suboptimal performance, indicating that purely class-driven prompts without image-specific context are less effective in guiding segmentation. Overall, BLIP provides a balanced captioning quality that is neither too generic nor overly specific, making it a suitable choice for guiding attention in our FA-Seg.


\begin{table}[!hb]
\caption{Performance of FA-Seg on the PASCAL VOC dataset using different methods for generating the text prompt $\mathcal{P}$.}
\label{tab:prompt}
\centering
\setlength{\tabcolsep}{9pt}
\begin{tabular}{lllccc}
\toprule
\textbf{Prompt type}         & \multicolumn{1}{c}{\textbf{}} &  &  &  & \textbf{mIoU (\%)} \\ \midrule
Simple text prompt           &                               &  &  &  & 63.6      \\
BIP \cite{blip1}             &                               &  &  &  & 64.0      \\
BIP2 \cite{blip2}            &                               &  &  &  & 64.0      \\
GIT \cite{GIT}               &                               &  &  &  & 63.7      \\
Gemini$^{(1)}$ \cite{gemini} &                               &  &  &  & 63.4      \\
Gemini$^{(2)}$ \cite{gemini} &                               &  &  &  & 63.7      \\ \bottomrule
\end{tabular}%
\end{table}

\subsubsection{Different pretrained model.}

We evaluated the FA-Seg framework using several pretrained diffusion models, including Stable Diffusion 1.5 and 2.1 \cite{sd}, as well as fast inference pretrained models such as Turbo SD \cite{Turbo}, SwiftBrush \cite{SwiftBrush}, SwiftBrush v2 \cite{SwiftBrushv2}, and 2-Rectified Flow (2-RF) \cite{instaflow}-the latter being the pretrained model adopted in our approach.

As shown in Table \ref{tab:pretrained}, when evaluated on the PASCAL VOC dataset, FA-Seg combined with Stable Diffusion 1.5 and 2.1 yields lower performance since these models are designed for multi-step image synthesis, whereas FA-Seg performs image inversion with minimal steps. In contrast, the fast inference models achieve higher segmentation accuracy, with 2-RF obtaining the best performance of 64\% mIoU. This superior performance can be attributed to the strong theoretical foundation of Rectified Flow \cite{rf1,rf2}, which reformulates the denoising process into a near-linear trajectory through flow matching, effectively simplifying a complex multi-step diffusion process into a smooth mapping from noise to data. Consequently, image inversion can be performed accurately with only one forward and one backward step, preserving semantic consistency without iterative optimization.

In comparison, methods such as SwiftBrush, SwiftBrush v2, and Turbo are primarily optimized for fast text-to-image generation and lack a well-grounded theoretical basis to guarantee semantically accurate reconstruction of input images. Although they enable faster inference, minor reconstruction errors can affect the attention maps, ultimately degrading the quality of the segmentation masks.

\section{Discussion and Conclusion}

\subsection{Limitations}

While FA-Seg achieves strong segmentation performance and inference efficiency, it still faces several limitations. First, the method is sensitive to the quality of the candidate class list, which must be provided or predicted in advance; incomplete or noisy class sets can lead to false negatives or misclassifications. Second, FA-Seg relies on static prompt encoding during inference-once the textual prompt is generated, it is not updated or refined adaptively, which may limit its flexibility in complex scenes. Third, as a training-free method, FA-Seg's performance is inherently constrained by the representational capacity of the pretrained diffusion backbone (e.g., Stable Diffusion); any limitations or biases in the underlying model directly affect segmentation quality and generalization.

\begin{figure}[]
	\centering 
	\includegraphics[width=0.5\textwidth]{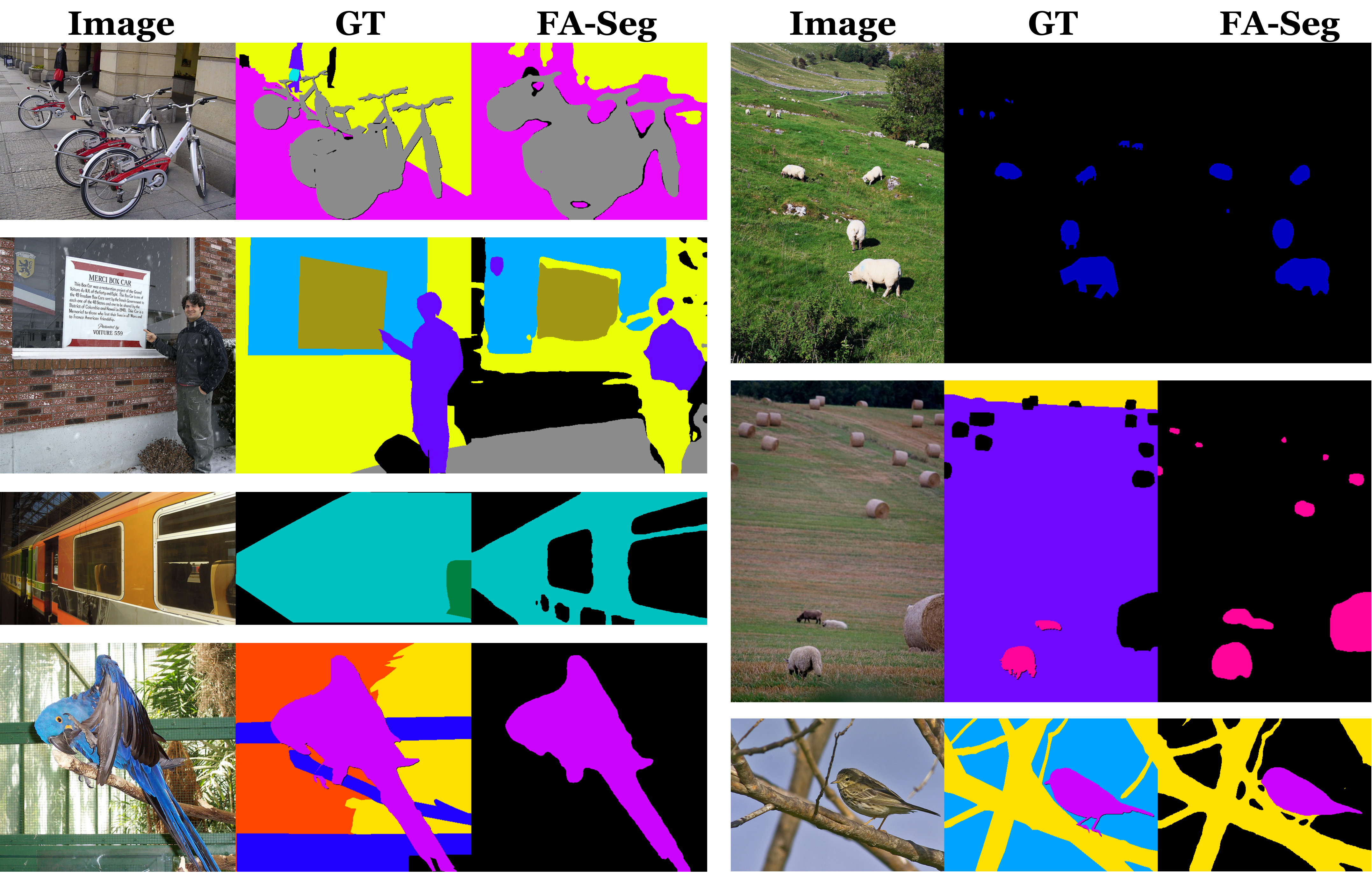}	
	\caption{Visualization of some failure cases by FA-Seg} 
	\label{fig:limit}%
\end{figure}

In addition, we present several failure cases of FA-Seg in Figure~\ref{fig:limit}, which highlight the current limitations of our approach in challenging segmentation scenarios. The first three rows in the left column illustrate difficulties in accurately segmenting object shapes and boundaries, particularly in scenes containing multiple overlapping objects (e.g., the first two rows). These cases suggest limitations in spatial discrimination when semantic boundaries are ambiguous or tightly packed. The top row of the right column demonstrates FA-Seg’s difficulty in detecting small objects, which may be overlooked due to their weak or missing representation in attention maps. Similarly, the second row of the right column shows segmentation errors when visually similar categories appear in close proximity, indicating challenges in category disambiguation under fine-grained visual similarity. Finally, the bottom row in both columns showcases cases where the quality of the candidate class list heavily influences segmentation performance. In these examples, the failure to predict or include correct class labels in the class prompt results in incomplete or incorrect segmentation, reaffirming FA-Seg’s dependency on accurate open-set label generation.

\begin{table}[]
\caption{Comparison of FA-Seg performance (mIoU\%) when used with different fast inference pretraineds.}
\label{tab:pretrained}
\centering
\setlength{\tabcolsep}{6pt}
\begin{tabular}{lllllllc}
\toprule
\textbf{Method}                     &  &  &  &  &  &  & \textbf{mIoU (\%)} \\ \midrule
Stable Diffusion 1.5 \cite{sd}                              &  &  &  &  &  &  & 58.9               \\
Stable Diffusion 2.1  \cite{sd}                    &  &  &  &  &  &  & 51.8               \\ \midrule
Turbo SD \cite{Turbo}                  &  &  &  &  &  &  & 60.0               \\
SwiftBrush \cite{SwiftBrush}        &  &  &  &  &  &  & 63.0               \\
SwiftBrush v2 \cite{SwiftBrushv2}     &  &  &  &  &  &  &  59.9                 \\

2 - Rectified Flow \cite{instaflow} &  &  &  &  &  &  & 64.0               \\ \bottomrule
\end{tabular}%
\end{table}

\subsection{Conclusion}
In this paper, we introduced FA-Seg, a novel training-free open-vocabulary semantic segmentation (OVSS) method that leverages diffusion models for efficient and accurate segmentation. Our method combines (1+1)-step DDIM inversion, a dual-prompt mechanism, and a multi-resolution attention fusion strategy. Unlike prior approaches, FA-Seg generates segmentation maps for all candidate classes in a single inference run, significantly reducing computational overhead. Additionally, we incorporate Test-Time Flipping (TTF) to improve prediction robustness. Extensive experiments on PASCAL VOC, PASCAL Context, and COCO Object benchmarks demonstrate that FA-Seg achieves state-of-the-art performance among training-free diffusion-based methods, with the highest average mIoU and achive inference time under one seconds. These results highlight the practicality and scalability of FA-Seg for real-world OVSS applications.

Building upon FA-Seg, future research can explore several directions to further enhance segmentation quality and flexibility, particularly by extending evaluations across diverse visual domains \cite{scene1,scene2,scene3,scene4,scene5,scene6,twin} and larger-scale datasets \cite{ADE20K,COCO-Stuff,bdd100k,lvis} to more comprehensively assess the model’s generalization capability. One promising avenue is the development of adaptive candidate class generation mechanisms, which leverage contextual cues to dynamically refine class sets and reduce dependency on the auxiliary candidate class generation process. Additionally, extending FA-Seg to support instance-level segmentation would enable more fine-grained visual understanding. With its high adaptability, accurate segmentation performance, and low segmentation latency, FA-Seg provides a practical and deployable solution for automatic segmentation in real-world applications.

\section{Acknowledgement}
This research was supported by The VNUHCM-University of Information Technology's Scientific Research Support Fund.

\printcredits

{\small
\bibliographystyle{IEEEtran}
\bibliography{ref}
}

\end{document}